\documentclass{article} 
\usepackage{GEM_workshop_2024,times}

\usepackage{amsmath,amsfonts,bm}

\def\eqref#1{equation~\ref{#1}}

\def\1{\bm{1}}

\def\vp{{\bm{p}}}

\DeclareMathAlphabet{\mathsfit}{\encodingdefault}{\sfdefault}{m}{sl}
\SetMathAlphabet{\mathsfit}{bold}{\encodingdefault}{\sfdefault}{bx}{n}

\usepackage{hyperref}
\usepackage{url}
\usepackage{caption}
\usepackage{subcaption}
\usepackage{graphicx}
\usepackage{tabularx, rotating, multirow}
\usepackage{makecell}
\usepackage{glossaries}
\usepackage[]{glossaries-extra}
\usepackage{relsize}
\usepackage{siunitx}

\setabbreviationstyle[acronym]{long-short}

\newacronym{ML}{ML}{machine learning}
\newacronym{MCTS}{MCTS}{Monte-Carlo tree search}
\newacronym{RL}{RL}{reinforcement learning}
\newacronym[longplural={Markov decision processes}, shortplural={MDPs}]{MDP}{MDP}{Markov decision process}
\newacronym{NVA}{NVA}{Neighbor Vector Algorithm}
\newacronym{NCA}{NCA}{Neighbor Count Algorithm}
\newacronym{SASA}{SASA}{Solvent Accessible Surface Area}
\newacronym{rSASA}{rSASA}{per Residue Solvent Accessible Surface Area}
\newacronym{MLP}{MLP}{multi-layer perceptron}
\newacronym{CNN}{CNN}{convolutional neural network}
\newacronym{LSTM}{LSTM}{long-short term memory network}
\newacronym{CPU}{CPU}{central processing unit}
\newacronym{UCT}{UCT}{upper confidence bound applied to trees}
\newacronym{CDF}{CDF}{cumulative distribution function}

\hypersetup{
    colorlinks=true,
    linkcolor=black,
    filecolor=black,      
    urlcolor=black,
    citecolor=black,
}

\title{Model-based reinforcement learning for protein backbone design}

\author{Frederic Renard $^{\dagger\mathsection}$\thanks{Work done during internship at InstaDeep Ltd} \And Cyprien Courtot \thanks{InstaDeep Ltd, 40B Rue du Faubourg Poissonnière, Paris, 75010} \And Alfredo Reichlin \thanks{Computer Science and Engineering, KTH Royal Institute of Technology,Stockholm, Sweden} \And Oliver Bent $^\dagger$\thanks{Corresponding authors: \{f.renard,o.bent\}@instadeep.com}}

\iclrfinalcopy 
\begin{document}

\maketitle

\begin{abstract}

Designing protein nanomaterials of predefined shape and characteristics has the potential to dramatically impact the medical industry. Machine learning (ML) has proven successful in protein design, reducing the need for expensive wet lab experiment rounds. However, challenges persist in efficiently exploring the protein fitness landscapes to identify optimal protein designs. In response, we propose the use of AlphaZero to generate protein backbones, meeting shape and structural scoring requirements. We extend an existing Monte Carlo tree search (MCTS) framework by incorporating a novel threshold-based reward and secondary objectives to improve design precision. This innovation considerably outperforms existing approaches, leading to protein backbones that better respect structural scores. The application of AlphaZero is novel in the context of protein backbone design and demonstrates promising performance. AlphaZero consistently surpasses baseline MCTS by more than 100\% in top-down protein design tasks. Additionally, our application of AlphaZero with secondary objectives uncovers further promising outcomes, indicating the potential of model-based reinforcement learning (RL) in navigating the intricate and nuanced aspects of protein design.

\end{abstract}

\section{Introduction}
\label{sec:intro}

The inverse design of proteins to optimise predetermined attributes is core to applications spanning from pharmaceutical drug development \citep{lagasse2017recent} to materials science \citep{dimarco2012multifunctional} or plastic recycling \citep{zhu2022enzyme}. \Gls{ML} has showcased its versatility in protein design, notably in the prediction of protein structures using AlphaFold \citep{jumper2021highly} and the design of protein sequences through ProteinMPNN \citep{dauparas2022robust}, significantly enhancing the capabilities of \textit{in silico} protein design.
Beyond structure prediction, \gls{ML} has proven to be highly effective in optimizing the complex and often irregular fitness functions associated with protein structures \citep{gront2012optimization, wu2019machine, gao2020deep}. Optimizing these fitness functions is particularly challenging due to the necessity of exploring the immense combinatorial space of amino acid sequences and structural configurations.

The success of \gls{RL} in complex combinatorial problems \citep{mazyavkina2021reinforcement}, such as the bin-packing \citep{laterre2018ranked} and traveling salesman \citep{khalil2017learning, grinsztajn2023winner} problems, underscores its potential in optimizing protein fitness functions. At its core, \gls{RL} operates on a simple yet powerful paradigm: an agent learns to make decisions by taking actions that maximize future rewards \citep{sutton2018reinforcement}. Model-based RL \citep{arulkumaran2017deep} differentiates itself by using models to simulate future states, allowing for strategic planning and foresight. This subfield of \gls{RL} was revolutionized by the introduction of the AlphaZero class of generalised game-playing \gls{RL} algorithms \citep{silver2016mastering, silver2017mastering, silver2018general}, which achieved state-of-the-art performance in the games of chess, shogi and go. AlphaZero navigates the vast tree of potential states and actions using a \gls{MCTS} guided by a policy-value neural network. DyNa-PPO \citep{angermueller2019model} pioneered the use of model-based \gls{RL} applied to protein design by modeling the design of proteins as a \gls{MDP} where amino-acid sequences are filled from left to right and the reward is chosen depending on the objective, such as optimizing the energy of protein contact Ising models \citep{marks2011protein} or transcription binding sites. EvoPlay \citep{wang2023self} later investigated the use of the single-player version of AlphaZero to design protein sequences and new luciferase variants. 

Focusing on the top-down design of protein nanomaterials of predefined shape, \citet{lutz2023top} developed a \gls{MCTS} approach to successfully design protein backbones while optimizing structural protein scores. This approach iteratively assembles protein secondary structures, alpha-helices and loops, to construct protein backbones. Cryo–electron microscopies \citep{bonomi2019determination} of the structures designed with this approach were almost identical to the \textit{in silico} designs. This novel use of \gls{MCTS} paves the way to investigate the application of AlphaZero on such a task.

We showcase the efficacy of AlphaZero in designing protein backbones of icosahedral shape while optimizing protein structural scores. Our contributions are threefold:
\begin{itemize}
    \item We benchmark AlphaZero against the \gls{MCTS} approach of \citet{lutz2023top} to compare performance in effectively sampling the protein backbone space to optimize the reward function.
    \item We demonstrate how the design of the \gls{MDP}, and more specifically of the reward function, can influence the learning and performance of AlphaZero.
    \item We propose a novel AlphaZero approach including side objectives to regularize the policy-value network throughout learning and benchmark it against the original AlphaZero algorithm.

\end{itemize}

\section{Methods}
\label{sec:meth}

\subsection{Markov Decision Process}

\paragraph{State and action spaces}
We formulate the protein backbone design problem as a \gls{MDP} $\mathcal{M} = (\mathcal{S}, \mathcal{A}, \mathbb{P}, r)$ with state space $\mathcal{S}$, action space $\mathcal{A}$, transition probabilities $\mathbb{P}$ and reward $r$. Both the state and action spaces follow the approach of \citet{lutz2023top}. The transition probabilities are \num{1} if the agent takes a legal action given its state, 0 else. The space of possible states is the ensemble of the protein backbones that fit inside of an icosahedron of predefined radius. The protein backbones are composed of alanines. Hence, each state is represented by a matrix of shape $( 5 \times \text{number of amino-acids}, 3)$ where every quintuple contains the cartesian coordinates of the three carbon atoms, the nitrogen atom and the oxygen atom of every alanine.
The space of possible actions is the union of three distinct subsets of action spaces; the actions of adding an alpha-helix of \num{9} to \num{22} residues on either end of the protein backbone; the actions of adding a loop sampled from one of \num{316} different loop clusters on either end of the protein backbone; or the action to terminate the episode. As described by \citet{lutz2023top}, beta strands are excluded from the design space to restrict the size of the design space. They could be added to generalize to a greater subset of proteins.


\paragraph{Reward} Adhering to the approach established by \citet{lutz2023top}, the reward is the combination of five different structural protein scores; the core score $C(s)$ quantifies the formation of an hydrophobic core; the helix score $H(s)$ quantifies if an alpha-helix is detaching from the rest of the protein backbone; the porosity score $P(s)$ assesses how porous the structure is; the monomer designability score $M(s)$ and the interface designability score $I(s)$ quantify how favorable the geometric interactions are, i.e. between core amino-acids and between the protein backbone's images by the icosahedral symmetries. The goal of the RL agents is to design protein backbones that meet score thresholds introduced by \citet{lutz2023top}, specifically: $C_0=0.2, ~H_0=2.0, ~P_0=0.45, ~M_0=0.9, ~I_0=17$. Combining those scores in a reward is a key step to ensure correct learning of the \gls{RL} agents. \\

Two different formulations of this reward are studied: \citet{lutz2023top} proposed a sigmoid reward formulation and we propose a novel thresholds reward formulation. The sigmoid reward is: 

\begin{equation}
    r(s) = \frac{m}{1 + \exp(-a (\Tilde{r}(s) - b))} ~~ \text{ with } \Tilde{r}(s) = 0.05 \prod_{S\in \{C, H, P, M, I \}} \sigma_S(S(s))
\label{eq:sigmoid_reward}
\end{equation}

With $m=1.0, a=0.03, b=200.0$, and where each score is normalized by the sigmoids $\sigma_S$. When trained with the sigmoid reward, algorithms will be noted \texttt{Algorithm} (sigmoid). 

The thresholds reward is :

\begin{equation}
    r(s) =  \frac{1}{\textstyle 5} \sum_S \max \left  ( \tau_0, \frac{S(s)}{S_0} \right ) \text{ with } S\in \{C, H, P, M, I \}
\label{eq:thresh_reward}
\end{equation}

Where $\tau_0$ is a hyperparameter. When trained with the thresholds reward, algorithms will be noted \texttt{Algorithm} (thresholds).  
In contrast to the sigmoid reward which pushes extremely small rewards to the sigmoid curve's tail, the thresholds reward reduces reward scarcity, setting $\tau_0$ as the target reward.

\paragraph{Episodes} Our methodology adopts the sequential design workflow for protein backbones introduced by \citet{lutz2023top} and is described in Figure \ref{fig:episode_fowchart}. First, an alpha-helix of five residues is initialized inside the icosahedron, uniformly sampling its position from a buffer of \num{5,000} initialization positions. Then, this alpha-helix is extended to a size between \num{9} to \num{22} residues on either the C-terminal or the N-terminal end. Next, an end is chosen and a loop is added on this end. Those last two steps are repeated until the terminal action is chosen, while enforcing that episodes can only terminate if the last secondary structure added was an alpha-helix. When the episode is finished, the reward is computed.
\\
Between each step, geometric checks are performed such that the secondary structure to be added cannot clash with the protein backbone or with the image of the protein backbone by one of its icosahedral symmetries. If the secondary structure does not meet these geometric conditions, the action is deemed illegal and cannot be taken by the agent.
\subsection{AlphaZero for protein backbone design}

\paragraph{AlphaZero algorithm}
\label{sec:original_az}
The original AlphaZero algorithm \citep{silver2018general} alternates between phases of self-play and phases of learning. In the self-play phases, episodes are completed by selecting at each step actions through a neural-network guided \gls{MCTS} search. At the end of each episode, the reward $r_T$,  the state $s_T$ and the tree policy $\boldsymbol{\pi}$ are stored in a buffer.
Then, during the learning phase, tuples $(\boldsymbol{\pi}, r_T, s_T)$ are sampled from the buffer and the policy and value estimates of the state $s_T$, $(\vp, v)$, are computed by the neural network. The policy $\vp$ is a vector of the probability over the actions given $s_T$ and the value $v$ is a prediction of the future reward of $s_T$. The parameters $\theta$ of the neural network are updated to minimize the loss :
\begin{equation}
    L_0 = (r_T - v)^2  - \boldsymbol{\pi}^T \log \vp + c||\theta||^2
\end{equation}

\begin{figure}[ht]
    \centering
    \includegraphics[width=0.63\textwidth]{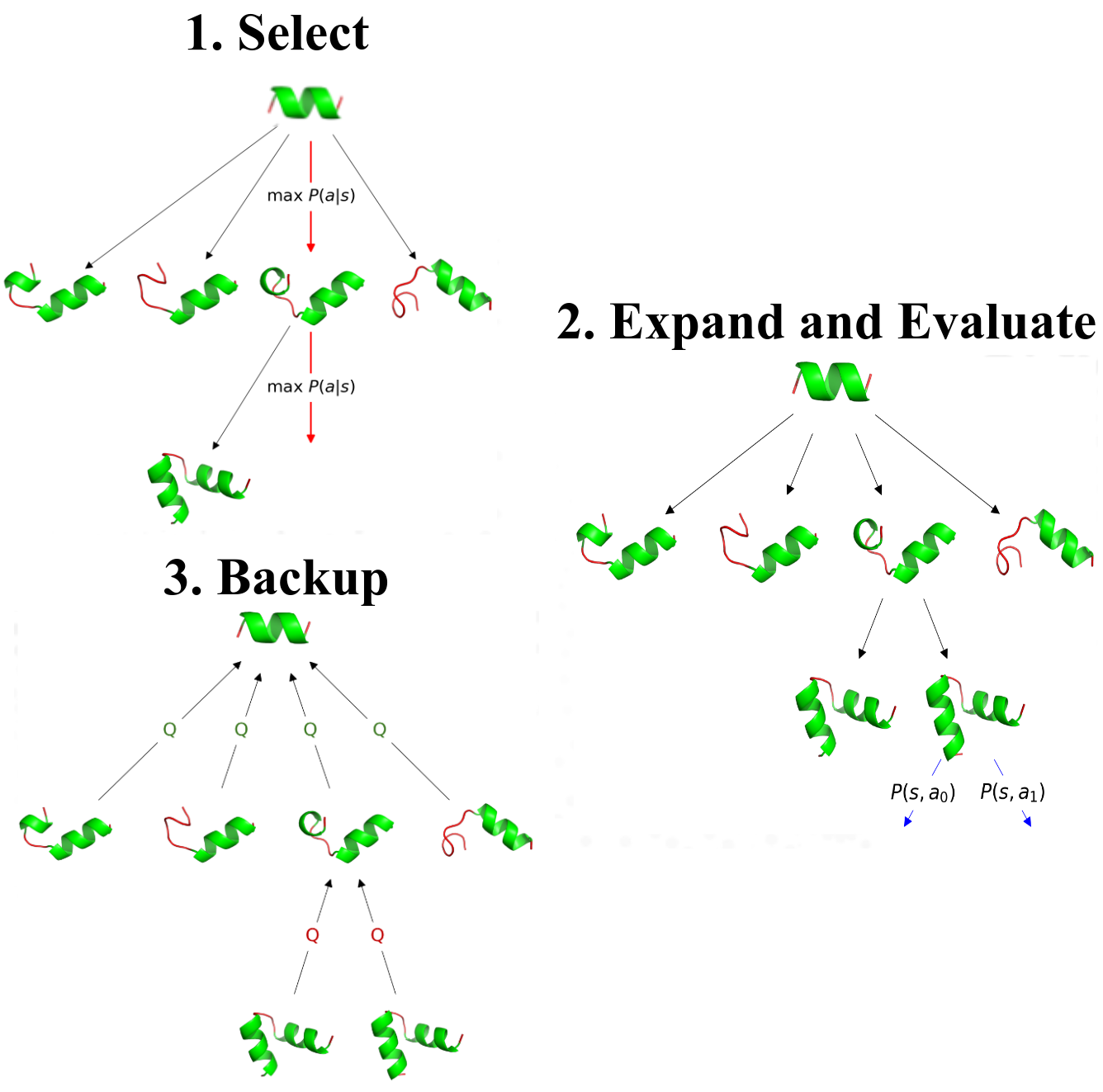}
    \caption{Diagram of the AlphaZero algorithm action selection process. Starting from the root node, the tree of states and actions is expanded by the repetition of the select, expand and evaluate, and backup phases. First, a new child node is selected by maximizing   $P(a|s) = Q(s, a) + c_{puct} P(s,a) \frac{\sqrt{\sum_b N(s, b)}}{1 + N(s, a)}$ with  $P(s,a)$ the policy network output, $Q(s,a)$ the mean action value of $(s,a)$ and $N(s,a)$ the number of visits of $(s,a)$. In the second phase, this new child node is evaluated by the neural network $f_\theta (s) = (P(s,a), V(s))$ with $V(s)$ the value network output. In the third phase, the value estimate $V(s)$ is used to update the $Q$ values for the parent nodes. After a number of \gls{MCTS} simulations, an action is selected according to $\boldsymbol{\pi}(a|s) = \frac{N(s,a)^{1/\tau}}{\sum_bN(s,b)^{1/\tau}}$. Once a terminal state is reached, $(\boldsymbol{\pi}, r_T, s_T)$ are stored in a buffer.}
    \label{fig:az_diagram}
\end{figure}

Figure \ref{fig:az_diagram} presents how the AlphaZero algorithm can be used to iteratively design protein backbones.

The most crucial hyperparameters for the AlphaZero algorithm are the exploration hyperparameters of the \gls{MCTS} search, typically noted $c_{puct}$ for the \gls{UCT} coefficient, $\tau$ which is a temperature parameter applied to the tree policy $\boldsymbol{\pi}$ to smoothen it and the number of \gls{MCTS} simulations performed at each step. In particular, the number of \gls{MCTS} simulations performed at each step impacts the diversity of the designs as a high number of \gls{MCTS} simulations allows the agent to explore more of the protein backbone space.

\paragraph{AlphaZero algorithm with side-objectives} In this paper, we propose to store for each terminal state, in addition to the previous elements, the value of the five scores for the protein backbone $s_T$ : $C(s_T), H(s_T),P(s_T), M(s_T), I(s_T)$.
The policy-value network is modified by adding five different heads which compute, given a protein backbone $s_T$, estimates for the five different scores: $\hat{C}(s_T), \hat{H}(s_T),\hat{P}(s_T), \hat{M}(s_T), \hat{I}(s_T)$.\\

\newpage
Then, during the learning phase, tuples $\left (\boldsymbol{\pi}, r_T, s_T, C(s_T), H(s_T),P(s_T), M(s_T), I(s_T) \right)$ are sampled from the buffer, the output by the neural network $(\vp, v, \hat{C}(s_T), \hat{H}(s_T),\hat{P}(s_T), \hat{M}(s_T), \hat{I}(s_T))$ of the state $s_T$ is computed and the parameters $\theta$ of the neural network are updated to minimize the loss :
\begin{equation}
    L = L_0 + \sum_{S\in \{C, H, P, M, I \}} \lambda_S(S(s_T) - \hat{S}(s_T))^2
\end{equation}

Where the $\lambda_S$ coefficients were chosen to scale the sum of the score losses to the order of magnitude of $\frac{L_0}{10}$ : $\lambda_C=1000, \lambda_H=1, \lambda_P=10, \lambda_I=0.1, \lambda_M=1$.
When comparing this AlphaZero approach to the original AlphaZero, it will be noted AlphaZero (side-objectives) as opposed to AlphaZero (original).

\subsection{Implementation}

The neural network architectures used for both AlphaZero algorithms are detailed in Appendix \ref{sec:az_archi}. Appendix \ref{sec:archi_search} shows the details of the architecture search that was performed to select expert networks for the protein structure scores.

\section{Results}

\subsection{Benchmark of MCTS against AlphaZero}

\begin{figure}[ht]
    \centering
    \includegraphics[width=0.95\textwidth]{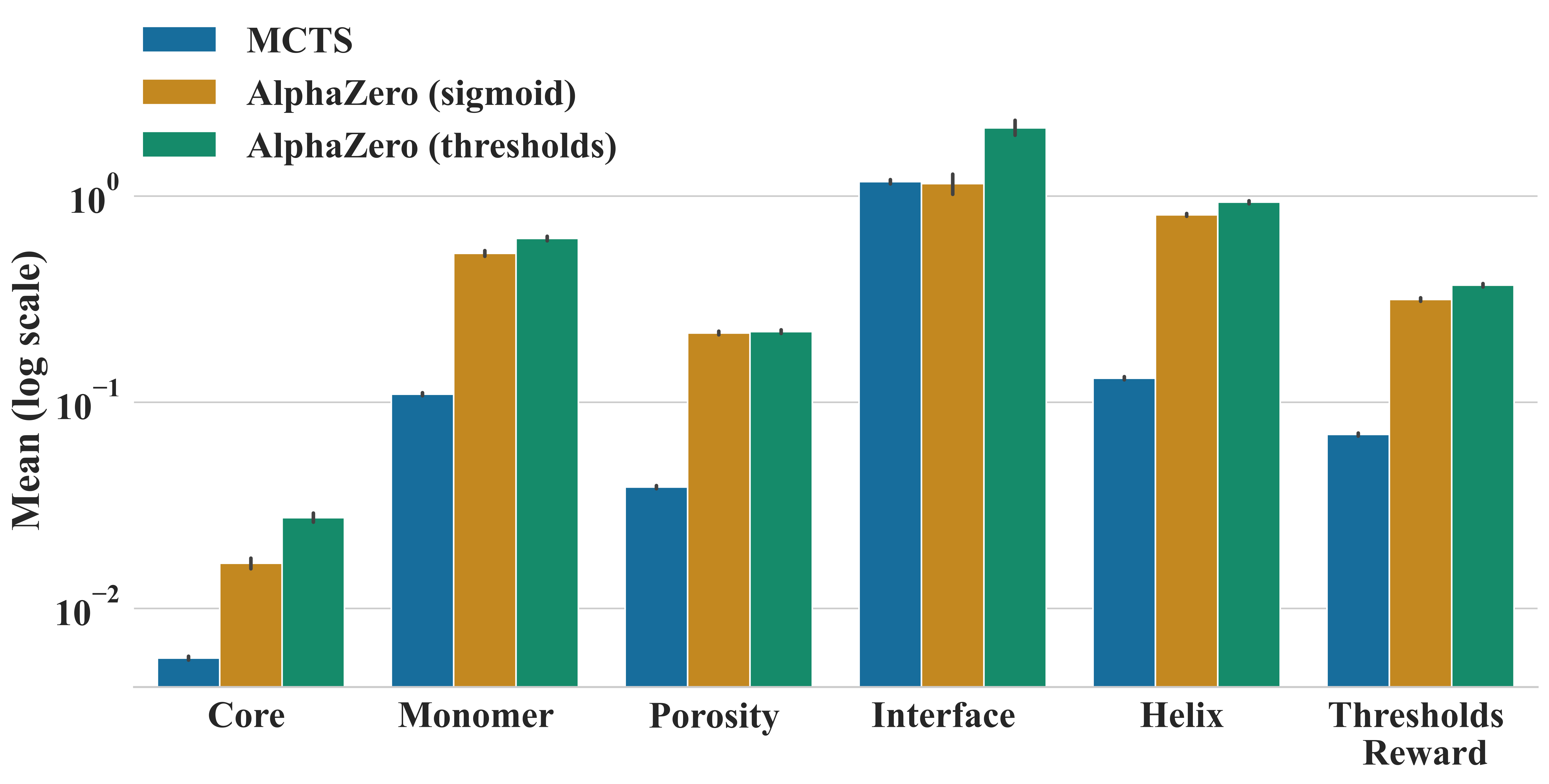}
    \caption{Protein score distributions means with \num{95}\% bootstrap confidence intervals. AlphaZero, and more specifically AlphaZero (thresholds) systematically outperforms \gls{MCTS} on all scores.}
    \label{fig:barplot_scores}
\end{figure}

\paragraph{Motivation and design}
The benchmark is designed to compare the score distributions of AlphaZero and MCTS with initializations inside the icosahedron that were not used to train AlphaZero. 100 alpha-helix initializations are generated inside an icosahedron of radius \num{75} Angstroms. \num{10,000} protein backbones are generated for each one of those initializations along with their scores with \gls{MCTS} using the code of \citet{lutz2023top}. Then, \num{300} protein backbones are generated with AlphaZero (sigmoid) and AlphaZero (thresholds) on each initialization with \num{300} \gls{MCTS} simulations at each step.
The hyperparameters used for training are referenced in Appendix \ref{sec:az_archi}, Table \ref{tab:az_hparams}. The metrics of interest are the score distributions of the protein backbones generated by the three methods.
The mean and \num{95}\% bootstrap confidence intervals for each score are summarized in Figure \ref{fig:barplot_scores}. 
\paragraph{Results}
First, we observe a clear superiority of both AlphaZero algorithms with respect to the \gls{MCTS} approach developed by \citet{lutz2023top}, with a mean improvement of factor \num{5} for the core score, \num{1.8} for the interface designability score, \num{7} for the helix score, \num{5} for the porosity score and \num{5} for the monomer designability score. The score distributions and p-values for the statistical tests are detailed in Appendix \ref{sec:benchmark_details}, Table \ref{tab:benchmark}. \\
Second, the p-values obtained when performing a Wilcoxon Rank-Sum test comparing AlphaZero (thresholds) to AlphaZero (sigmoid), reported in Appendix \ref{sec:az_archi}, indicate that AlphaZero (thresholds) consistently reaches better performance compared to AlphaZero (sigmoid). 

\subsection{Benchmark of AlphaZero with and without side-objectives}

\begin{figure}[ht]
    \centering
 \includegraphics[width=0.9\textwidth]{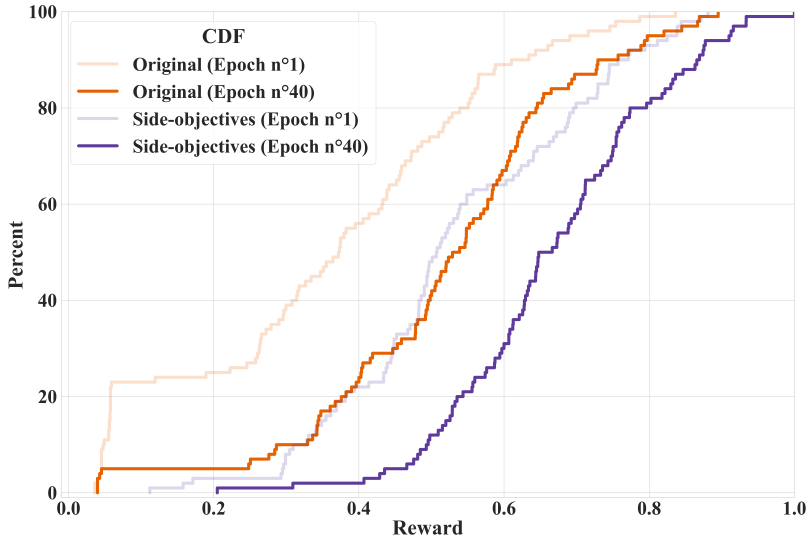}
        \caption{\gls{CDF} of the reward of both algorithms at the first epoch and at epoch \num{40} of training. AlphaZero (side-objectives) consistently achieves higher rewards compared to the AlphaZero (original).}
    \label{fig:cumul_rwd}
\end{figure}

\paragraph{Motivation and design}
To compare both algorithms, the reward distributions at each step of training are collected for 50 epochs.  Training hyperparameters are referenced in Appendix \ref{sec:az_archi}, Table \ref{tab:az_hparams}.
Both algorithms are compared with the thresholds reward formulation. The parameter $\tau_0$ of Equation \ref{eq:thresh_reward} is set to \num{1}. 
The metrics of interest are the \glspl{CDF} of the rewards at epoch 1 and epoch 40, presented in Figure \ref{fig:cumul_rwd}. The \glspl{CDF} are computed on batches of 55 to 65 episodes that are self-played by the AlphaZero agent before each learning phase as describedin Section \ref{sec:original_az}. 

\paragraph{Results} The \gls{CDF} of AlphaZero (side-objectives) is consistently to the right compared to the \gls{CDF} of the original AlphaZero, achieving higher rewards. The maximum reward per batch of episodes of AlphaZero (side-objectives) reaches \num{1.0} for \num{28}\% of the epochs, which indicates that all five objectives are simultaneously achieved. The average reward of AlphaZero (side-objectives) is consistently above to the mean reward of the AlphaZero (original). The mean episode reward and maximum episode reward of both algorithms throughout the training are reported in Appendix \ref{sec:az_archi} in Figure \ref{fig:rwd_training}.

\section{Discussion}

In this work, we have demonstrated the pertinence of model-based deep \gls{RL}, and more specifically of AlphaZero, for protein backbone design. Both the impact of the reward formulation and the addition of side-objectives emerge as crucial elements for achieving all objectives with AlphaZero. In comparison with the designability, diversity and novelty metrics of diffusion models \citep{yim2023se}, the designability is evaluated by the different scores and the diversity and novelties are constrained by the different protein secondary structures that can be chosen, as described in Appendix \ref{sec:prot_struct}.

Several improvements can be made to this work. First, considering the superiority of the AlphaZero agent trained with the threshold reward compared to the sigmoid reward, shaping the reward function is key to achieve better performance. Scheduling the increase of the parameter $\tau_0$ throughout training could allow the agent to learn how to outperform the score thresholds through a curriculum learning approach \citep{narvekar2020curriculum}. However, if such an approach is taken, the performance of the agent in the five different scores should be monitored to ensure the agent does not overspecialize in one score, hurting its performance with the other scores. Another possible improvement would be to search for better possible expert networks architectures for the AlphaZero algorithm with side-objectives. Those side-objectives regularize the policy-value network, improving its performance in designing protein backbones. Future work could also investigate the use of AlphaZero to generate protein backbones of other shapes such as nanopores as in the work of \citep{lutz2023top} and the transfer learning capabilities of AlphaZero agents. The approach could be enriched by designing protein sequences from the protein backbones with the methodology of \citet{lutz2023top} and employing AlphaFold for a structure prediction test of the designs obtained.

 This work paves the way for the use of \gls{RL} in multi-objective optimization of protein structures. The application of AlphaZero to generate protein backbones alleviates some of the common drawbacks of machine learning, as every design and choices that were made are traceable. It unlocks new methods to design protein nanomaterials of specific shapes to target therapeutic sites of interest or innovative materials. 

\clearpage

\bibliography{iclr2024_conference}
\bibliographystyle{iclr2024_conference}

\clearpage

\appendix
\section{Appendix}

\subsection{Benchmark of MCTS against AlphaZero details}
\label{sec:benchmark_details}

 Table \ref{tab:benchmark} presents the mean and standard deviation of the distributions for all five scores for the protein backbones generated by the algorithms, and for each score the p-value of the Wilcoxon Rank-Sum test with the null hypothesis being :  the mean of the scores collected by the AlphaZero agent is greater than the mean of the scores collected by the \gls{MCTS} agent.\\
In this table, means are noted with $\mu$, standard deviations with $\sigma$ and the p-values of the Wilcoxon Rank-Sum test with $p$. P-values below \num{1e-308} are rounded to zero.\\

\begin{table}[ht]
\caption{Benchmark of MCTS against AlphaZero agents.}
\label{tab:benchmark}
\centering
\begin{tabularx}{\textwidth}{*{4}{>{\centering\arraybackslash}X}}
\hline
& \thead{\bf MCTS baseline } 
&\thead{\bf Original AlphaZero \\ \bf with \\ \bf sigmoid reward} 
&\thead{\bf Original AlphaZero \\ \bf with \\ \bf thresholds reward} \\ 
\hline
$\boldsymbol{\mu_C}$ & \num{0.0056} &  \num{0.017} & \num{0.028}\\
$\boldsymbol{\sigma_C}$ & \num{0.02}  & \num{0.027} & \num{0.038} \\
$\boldsymbol{p_C}$ & - & \num{2.2e-117}  & 2\num{2.1e-274} \\
\hline
$\boldsymbol{\mu_H}$ & \num{0.13} & \num{0.81} & \num{0.93} \\
$\boldsymbol{\sigma_H}$ & \num{0.38}  & \num{0.36} & \num{0.43} \\
$\boldsymbol{p_H}$ & - & \num{0}  & \num{0} \\
\hline
$\boldsymbol{\mu_P}$ & \num{0.0390}  & \num{0.216} & \num{0.220} \\
$\boldsymbol{\sigma_P}$ & \num{0.121} & \num{0.117} & \num{0.123} \\
$\boldsymbol{p_P}$ & - & \num{0} & \num{0}\\
\hline
$\boldsymbol{\mu_I}$ & \num{1.17}  & \num{1.15} & \num{2.14}  \\
$\boldsymbol{\sigma_I}$ & \num{4.64}  & \num{3.81} & \num{5.21} \\
$\boldsymbol{p_I}$ & - & \num{7.0e-4} & \num{6.14e-25} \\
\hline
$\boldsymbol{\mu_M}$ & \num{0.11}  & \num{0.53} & \num{0.62} \\
$\boldsymbol{\sigma_M}$ & \num{0.30}  & \num{0.47} & \num{0.46} \\
$\boldsymbol{p_M}$ & - & \num{0} & \num{0} \\
\hline
\end{tabularx}

\end{table}

\subsection{AlphaZero algorithm}
\label{sec:az_archi}

The neural network architecture used for AlphaZero (original) is a \gls{MLP} with shared parameters for the value and policy networks. The input of this network is the flattened protein backbone array. Then, two hidden layers with \num{512} and \num{256} neurons were added before two different two-layers heads: the policy head with output size of \num{663} and the value head with output size \num{1}. LeakyRelu \citep{he2015delving} activation with a negative slope of \num{0.01} were used between each layers.

\begin{figure}[ht]
\centering
    \includegraphics[width=\textwidth]{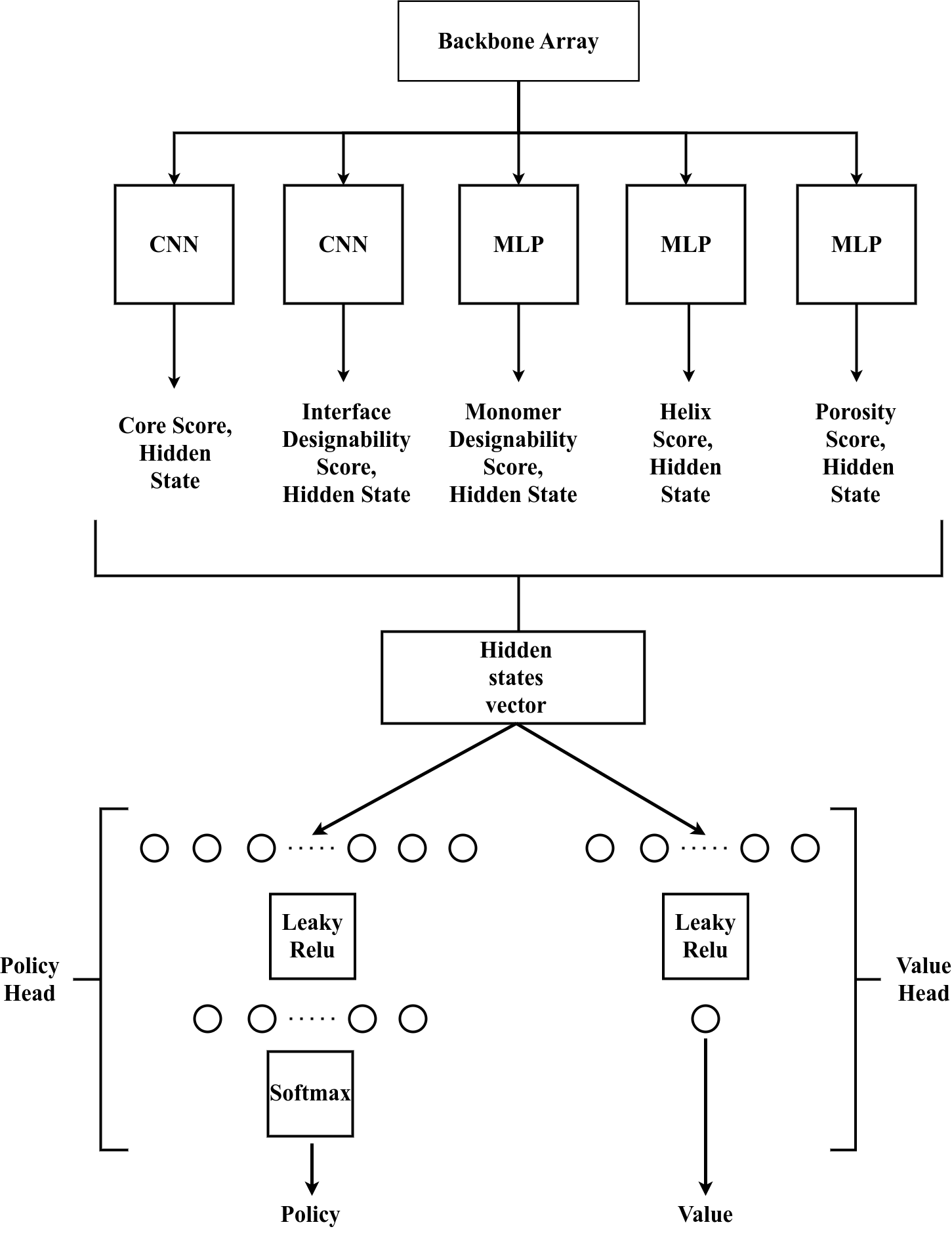}
    \caption{Mixture of experts architecture for AlphaZero. Circles represent linear layers. \Glspl{CNN} are used to predict the core and interface designability score and \glspl{MLP} for the other scores. The hidden states used to compute the scores are concatenated and used by two different heads : a policy and a value head. The neural network output the policy, the value and the five different protein structure scores.}
    \label{fig:az_mixture}
\end{figure}

The mixture of experts architecture used for AlphaZero (side-objectives) is shown in Figure \ref{fig:az_mixture}. The \gls{CNN} and \gls{MLP} used are identical to the ones defined in section \ref{sec:archi_search}.
The hyperparameters used for the training of AlphaZero are presented in Table \ref{tab:az_hparams}. 
\gls{RL} experiments were performed with \num{19} workers collecting rollouts and one worker performing the learning steps. Each worker used an AMD EPYC \num{7452} CPU with \num{5}GB of RAM. 

When performing a Wilcoxon Rank-Sum test with the null hypothesis being : the mean of the scores collected by AlphaZero (thresholds) is greater than the mean of the scores collected by AlphaZero (sigmoid), the p-values obtained are : \num{6.43e-24} for the core score, \num{4.90e-24} for the helix score, \num{1.31e-16} for the monomer designability score, \num{8.89e-8} for the interface designability score and \num{0.045} for the porosity score.

\begin{figure}[ht]
    \centering
 \includegraphics[width=\textwidth]{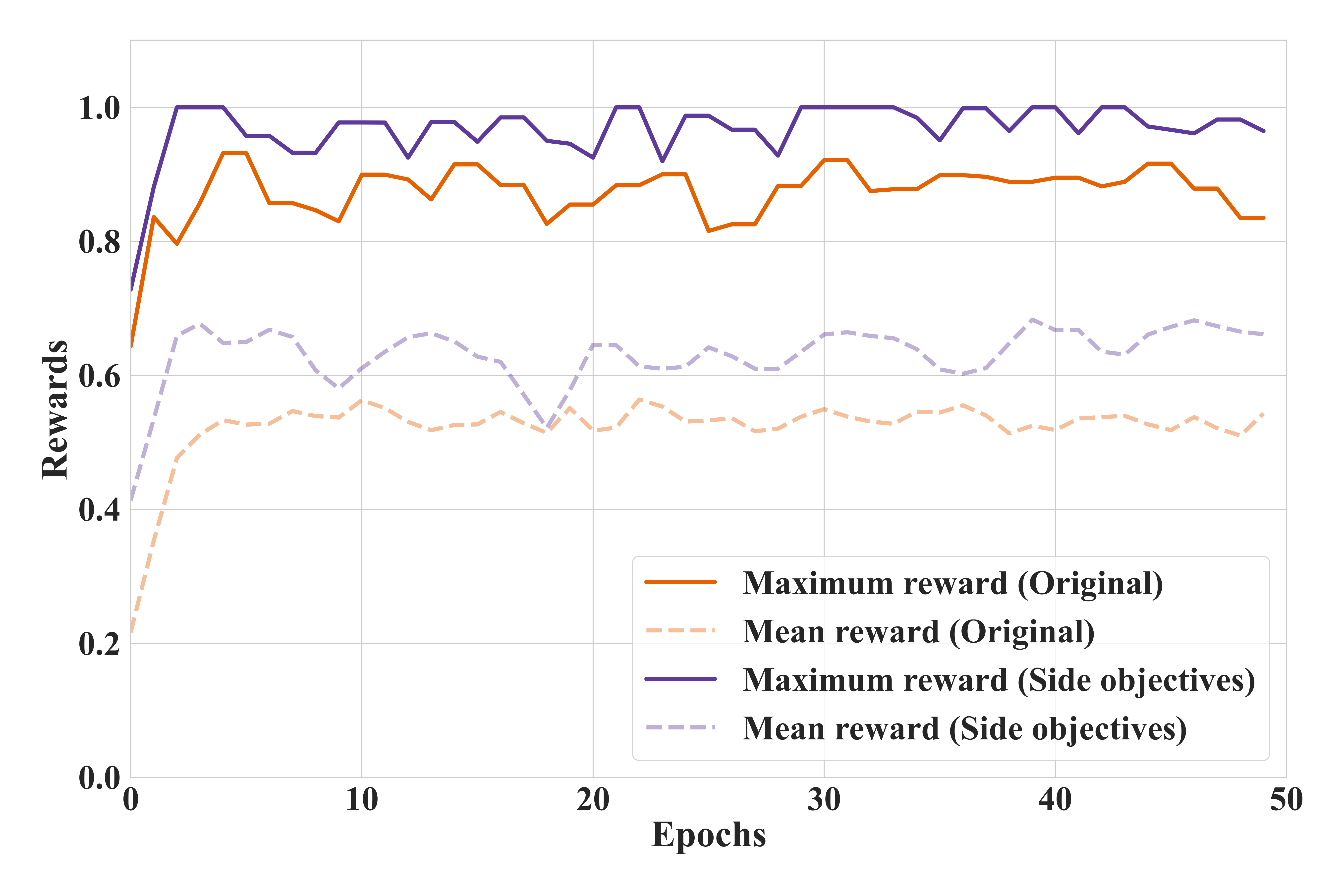}
        \caption{Rewards of the AlphaZero agents throughout training. Both AlphaZero (side-objectives) maximum and mean rewards are consistently higher than those of AlphaZero (original).}
    \label{fig:rwd_training}
\end{figure}

The mean episode reward and maximum episode reward of both agents is reported in Figure \ref{fig:rwd_training}. Those reward statistics are computed on batches of \num{55} to \num{65} episodes that are self-played by the AlphaZero agent before each learning phase as described in Section \ref{sec:original_az}. 
Both agents quickly converge and AlphaZero (side-objective) consistently reaches higher reward statistics. Training lasted \num{5}h for the AlphaZero (original) and \num{7}h for AlphaZero (side-objectives).

\begin{table}[ht]
\caption{Hyperparameters for training the AlphaZero agents.}
\label{tab:az_hparams}
\centering
\begin{tabular}{ll}
\hline
\textbf{Hyperparameter} & \textbf{Value} \\ \hline
train batch size & \num{1024} \\ 
learning rate & \num{5e-5} \\ 
L2 regularization coefficient & \num{1e-5} \\ 
$c_{puct}$ & \num{1.5} \\ 
$\tau$ & \num{1.0} \\ 
Number of MCTS simulations & \num{5000} \\ 
Buffer size & \num{100,000} \\ 
Number of training iterations & \num{200} \\ 

\end{tabular}
\end{table}

\clearpage
\subsection{Neural networks architecture search}
\label{sec:archi_search}

In order to determine what neural network architecture to use for the AlphaZero algorithm,  \num{80,000} protein backbones, represented by matrices of shape $(400, 3)$ were generated through an \gls{MCTS} search with their corresponding scores. Then, several architectures were tested on the supervised learning task of predicting the scores from the protein backbones, including a \gls{MLP}, a \gls{LSTM} \citep{yu2019review} and a mixture of experts composed of two one-dimensional \glspl{CNN} for the core and interface designability score and three \glspl{MLP} for the other scores. This dataset is augmented by computing the image for each backbone by 8 icosahedral symmetries, yielding a dataset of \num{640,000} protein backbones. It is split between a train dataset of size \num{512,000}, a validation dataset of size \num{80,000} and a test dataset of size \num{80,000}. Datasets are standardized using the train dataset mean and standard deviation.
Neural networks are trained to minimize the Mean Squared Error with the training hyperparameters shown in Table \ref{tab:train_hparams}. \\
The \gls{MLP} architecture takes as input the flattened backbone array of shape \num{1200} and outputs a vector of shape \num{5}. It is a three layers network with respectively \num{512}, \num{256} and \num{5} neurons in each layer and LeakyReLu activation functions with a negative slope of \num{0.01} after each layer, including the output layer.\\
The \gls{LSTM} architecture has the same inputs and outputs as the \gls{MLP}. It is composed of two stacked \gls{LSTM} layers with a hidden size of \num{128}, followed by a linear layer of size \num{64}, LeakyRelu activation with a negative slope of \num{0.01} and a final linear layer of size \num{5}.\\
The \gls{CNN} architecture accepts as an input backbone arrays of shape (\num{400}, \num{3}) and outputs a vector of size \num{1}. LeakyReLu activation functions were used after each layer with a negative slope of \num{0.01}, including the output layer. Its hyperparameters are presented in  Table \ref{tab:convnet_hparam}.\\

\begin{table}[ht]
\caption{\Gls{CNN} hyperparameters.}
\centering
\begin{tabular}{cc}
\hline
\textbf{Hyperparameter} & \textbf{Value} \\ \hline
Number of channels first convolutional layer & \num{3} \\ \
Kernel size first convolutional layer & \num{5} \\ \
Stride first convolutional layer & \num{3} \\ \
Number of channels second convolutional layer & \num{16} \\ \
Kernel size second convolutional layer & \num{2} \\ \
Stride second convolutional layer & \num{2} \\ \
Size of first fully connected layer & \num{528} \\ \
Size of second fully connected layer & \num{64} \\ \
Size of third fully connected layer & \num{32} \\ \

\end{tabular}
\label{tab:convnet_hparam}\\
\end{table}

All networks are trained with the Adam algorithm \citep{kingma2014adam} and a Cosine Annealing learning rate schedule \citep{loshchilov2016sgdr} to minimize the Mean Squared Error loss between the scores and the network's predictions. For the mixture of experts, the networks are trained with different learning rates.
The parameters used for training are shown in Table \ref{tab:train_hparams}. \\
Supervised learning experiments were performed on a \num{12}th Gen Intel® Core™ i7-1265U × 12 \gls{CPU} with \num{32} Gio of RAM.

\begin{table}[h]
\centering
\caption{Training hyperparameters for the architecture search.}
\begin{tabular}{cc}
\hline
\textbf{Hyperparameter} & \textbf{Value} \\ \hline
MLP Learning Rate & \num{5e-3}\\ \
LSTM Learning Rate & \num{5e-3} \\ \
Core score CNN Learning Rate & \num{5e-4} \\ \
Interface Designability score CNN Learning Rate & \num{1e-3} \\ \ 
Monomer Designability score MLP Learning Rate & \num{1e-3} \\ \
Helix score MLP Learning Rate & \num{5e-3} \\ \
Porosity score MLP Learning Rate & \num{1e-3} \\ \
Minimum learning rate of cosine annealing & \num{5e-6} \\ \
Minimum learning rate reached at epoch & \num{100} \\ \
Number of training epochs & \num{100} \\ \
Adam weight decay & \num{1e-5} \\ \
Batch size & \num{1024} \\ \
\end{tabular}
\label{tab:train_hparams}
\end{table}

\begin{figure}[ht]
\centering
    \includegraphics[width=\textwidth]{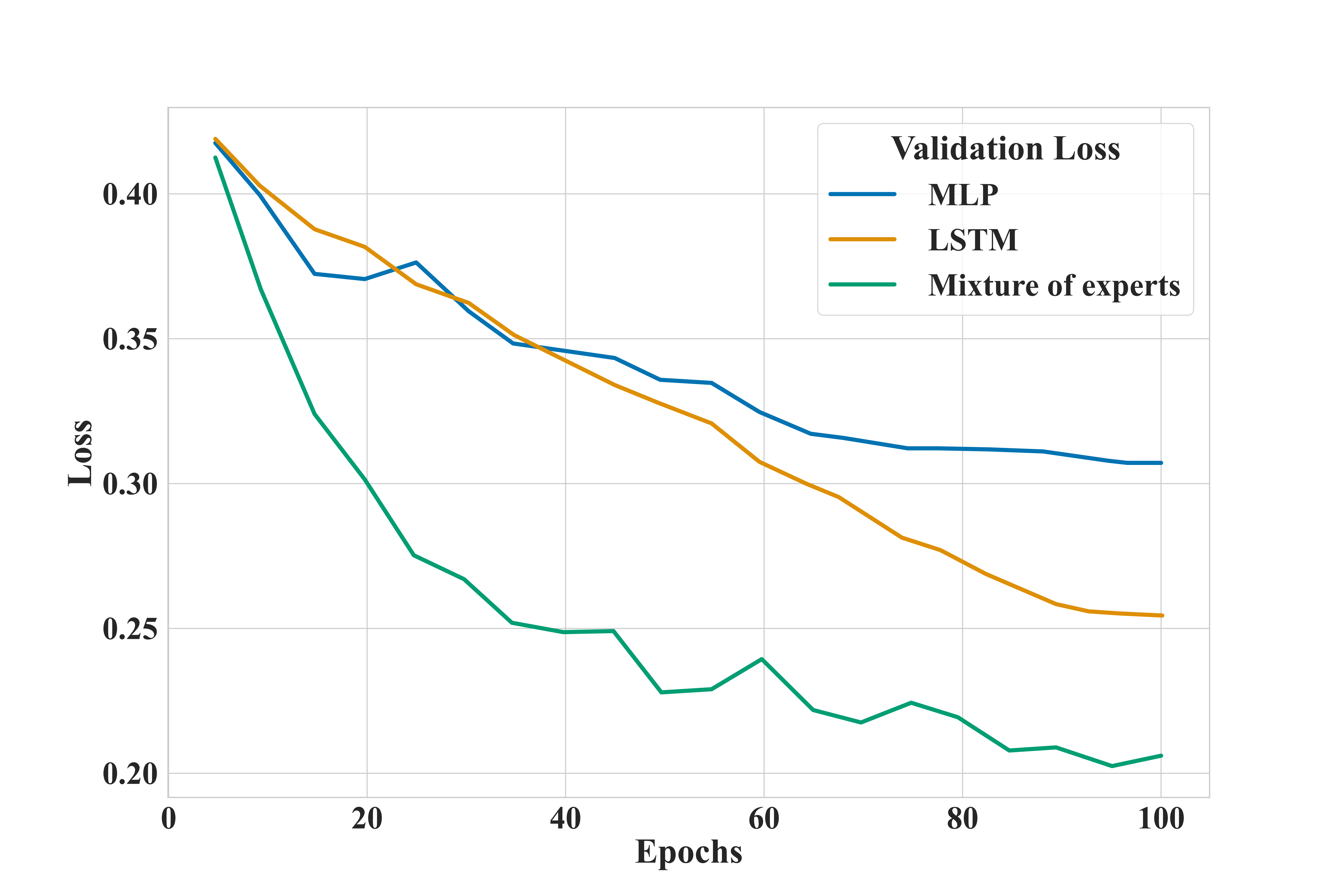}
    \caption{Validation losses for the prediction of protein structure scores from protein backbones. Results show a clear superiority of the mixture of experts network both in terms of speed of convergence and validation loss.}
    \label{fig:architecture_loss}
\end{figure}

Figure \ref{fig:architecture_loss} presents the validation losses for all three tested architectures. The mixture of experts proves to be the best architecture both in terms of speed of convergence and validation loss.

\clearpage

\subsection{Protein Backbone Design Episode}

\begin{figure}[ht]
\centering
    \includegraphics[width=0.51\textwidth]{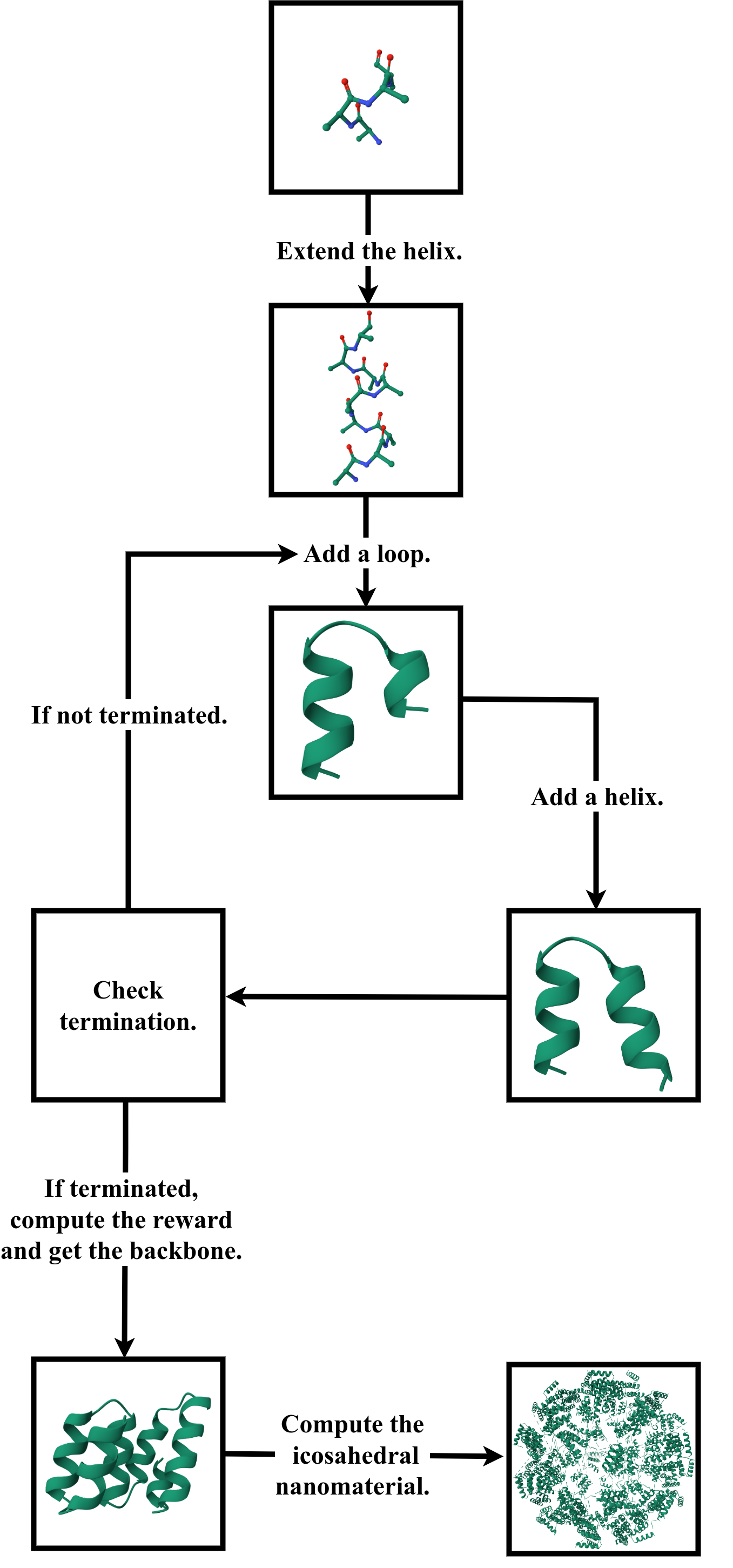}
    \caption{Flowchart of an episode. During an episode, alpha-helix of between 9 to 22 residues and loops sampled from 316 different loop clusters are iteratively added to the protein backbone. Between each step, geometric checks are performed to avoid clashes between the structure to be added and the current protein backbone or one of its icosahedral symmetries. Geometric checks also prevent the protein backbone from exiting the icosahedral shape. If an action passes the geometric checks, it is legal. If no legal actions can be found, the episode is ended. If the episode ends because no legal actions can be found and the last action was to add a loop, this loop is removed. An episode is terminal if more than 7 alpha-helices were added, if the number of amino-acids is superior to 80 or if the terminal action was chosen. The terminal action can be chosen if the backbone has more than 3 alpha-helices and is always chosen if no legal actions can be found. At each helix-addition step, all legal helix action additions can be chosen. At each loop-addition step, a subset of 50 loop clusters is randomly chosen and one loop from each cluster is sampled. The loop actions the agent can take are the legal loop actions of this subset of loop actions.}
    \label{fig:episode_fowchart}
\end{figure}

\clearpage

\subsection{Protein Secondary Structures}
\label{sec:prot_struct}
\begin{figure}[ht]
    \centering
\includegraphics[width=0.49\textwidth]{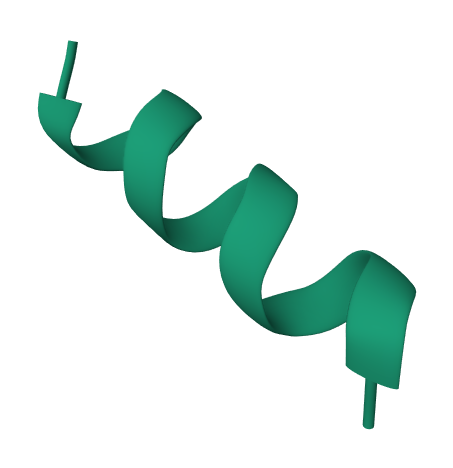}
    \caption{Example of alpha-helix used to construct protein backbones. Image generated with Mol* \citep{Mol}.}
    \label{fig:helix}
\end{figure}

\begin{figure}[ht]
    \centering
    \begin{subfigure}{0.49\textwidth}
        \includegraphics[width=\textwidth]{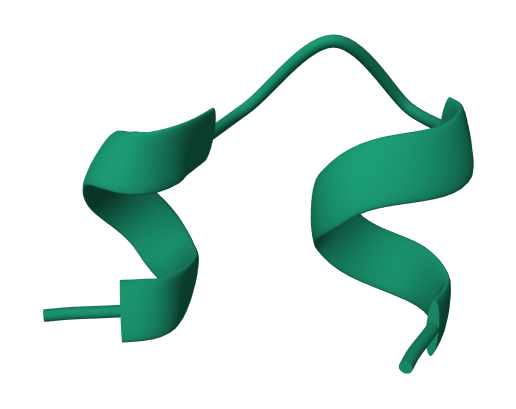}
    \end{subfigure}
    \includegraphics[width=0.49\textwidth]{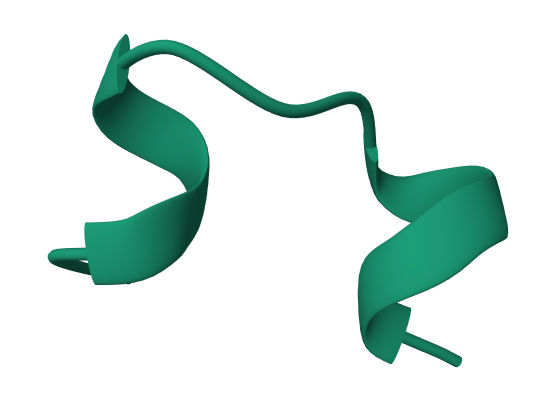}
    \caption{Examples of protein loops used to construct protein backbones. Images generated with Mol* \citep{Mol}.}
    \label{fig:loops}
\end{figure}

This section describes the different protein secondary structures used to construct protein backbones. Figure \ref{fig:helix} presents an example of alpha-helix used to construct the protein backbones and Figure \ref{fig:loops} presents two different protein loops, amongst more than 20.000, that can be chosen by the agent. The different orientations given to the alpha-helices by the protein loops increase the diversity of the generated backbones while ensuring designability by optimizing the different protein structure scores.

\end{document}